# New Methods of Analysis of Narrative and Semantics in Support of Interactivity

Fionn Murtagh (1,2), Adam Ganz (3) and Joe Reddington (1)

(1) Department of Computer Science
(2) Science Foundation Ireland
(3) Department of Media Arts
Royal Holloway, University of London
Contact: fmurtagh@acm.org

**Abstract**

Our work has focused on support for film or television scriptwriting. Since this involves potentially varied story-lines, we note the implicit or latent support for interactivity. Furthermore the film, television, games, publishing and other sectors are converging, so that cross-over and re-use of one form of product in another of these sectors is ever more common. Technically our work has been largely based on mathematical algorithms for data clustering and display. Operationally, we also discuss how our algorithms can support collective, distributed problem-solving.

## 1. Introduction

Papadimitriou (2003) succinctly provides a wide range of examples of where storytelling is or should be central in teaching. He ends with an example from author Umberto Eco, termed "salgarism" by the latter, the "telltale symptoms" of which are "incongruity and discontinuity between story and embedded information". In our work we seek to find algorithmically such patterns or trends in the data. We do this, it must be noted, not just on a word level, or a sentence level, or a paragraph level, but rather in close association with some level or levels of information resolution. Regarding the latter (level of information resolution), it will be observed in this article that it is based on clusters of words, which may or may not be taking order or sequence of word placement into account, and, secondly, which are typically based on a hierarchical clustering so that resolution scale in this context can be associated with level in such a hierarchy.

When one considers multiple platforms, encompassing media, format, and form and extent of interactivity, Murray (1997) raises the question of what platform and/or delivery mechanism is superior. In answering this, she points to our often very simplistic understanding of story. Instead, to explore commonality across delivery platform, it is necessary, she notes, to capture meaning, i.e. semantics.

We will next look at recent work in the area of games-related authoring software, which is also a prime objective of our work.

More traditional forms of authoring software have been largely focused on identifying and using interactions between, characters, events, authors and the like (see Louchart et al., 2008). More recent approaches to narrative synthesis include Louchart et al. (2008) for whom there is a "narrative paradox" in the imputed distance between plot and interaction, insofar as plot implies abstracting away from, and in a sense pulling against, the interactive environment.  This leads to a focus on process of creating a plot narrative, rather than the structure of the story.  As Louchart et al. (2008) note: "In EN [emergent narrative] we try to remove the need to 'think in terms of plot', because the notion of plot … has a problematic tension with the role of the interactor."  In our work, instead, we focus precisely on the structure of narrative, and show how readily we can reconcile this approach with an interactive environment (cf. section 5 below).  Kriegel and Aylett (2008) refer to bottom-up collaborative authoring based on such an approach as remaining "incoherent and chaotic", which is not the case with our approach.

Aylett's emergent narrative (EN) approach is developed too by Swartjes and Theune (2009a), which is based on branching and causal networks in the context of an "overall space of possible stories".  Authors can modify or accept the choice of story elements and prolongations proposed by a story generator (referred to as, respectively, debugging or co-creation).  In concluding, Swartjes and Theune (2009a) question whether their own approach is capable of scaling.  Further work in the symbolic artificial intelligence tradition can be seen in Swartjes and Theune (2009b). In our approach, on the other hand, our main possible limitation on scaling is due to (computationally cubic order) eigen-reduction, which, because this is carried out on very sparse matrices, is unlikely to be a limitation (for computational work on large sparse matrices, see e.g. Murtagh et al., 2000).

In this article, rather than looking at narrative applied specifically to games we are looking to use algorithmic techniques, that are also very relevant for games, to find and depict the narrative thread in screenplay, by defining the times when a screenplay becomes most like and most different from itself.  We use screenplay as our testbed or environment.

A so-called "linear" story consists of conflicts or moments of decision, and it is our contention that using textual narrative to identify the deep structure of a screenplay will be extremely helpful in thinking about the development of serious games.  A screenplay in fact is written to maximize the number and impact of those nodal narrative points which are, or will be, experienced by the audience in real time.

 A problem of the tree structure model prevalent in game narratology (cf. discussion above, e.g. Swartjes and Theune, 2009a, 2009b), is that it does not recognize that the experience of any audience member will be to experience their individual journey as a single pathway regardless of how many possible pathways they might have taken.

Indeed the notion of foregrounding story over plot is not just a function of the interactive environment. Stephen King, the novelist, whose work has inspired more successful original films than perhaps any other writer working today, has written that "plotting and the spontaneity of real creation aren't compatible" (King, 2002: 164).  Plot, he calls "a dullard's first choice.  The story which results from it is apt to feel artificial and labored".  So instead of starting with plot, King puts his characters in a situation and likes to "watch them try to work themselves free".

The dramatic narrative, including a novel or filmscript, is devised to model those moments of choice, and to involve the audience as an active collaborator in imagining the story, which involves *all* the future outcomes desired and feared by the audience and not just the narrative choices which actually occur in the script or film. See Ganz (2010) for further discussion of this.

By developing a toolset or processing environment, that allows narratives to be represented numerically, we are finding the possible patterns of narrative desired or created by the audience. Some of this work has been accomplished up to now by measuring reaction but it has not been done by analyzing the text itself and using these analysis outcomes to create heuristics for narrative in a form can be recognized by a machine. These heuristics could also well create parameters for narrative in serious games and, for example, be subsequently used to close down narrative options at certain stages of a game.

## 2. Analysis and Synthesis, Episodization and Narrativization

We have explored film and TV scripts in the converging world of cinema, television, games, online or virtual societies, with applications in entertainment, education and many other areas. In this context, our primary focus has been on the tracking of narrative. Further new application domains, and our work to date in these domains, are reviewed in this article.

We explore (i) features or attributes of narrative that we can determine and measure, based on semantic analysis, using the movie Casablanca; (ii) segmentation – "episodization" is the term used in Aristotle's Poetics – and its use, illustrated by cases from the CSI Las Vegas television series; and (iii) narrative synthesis, including both a new technical approach to this, and our experience with a collaborative narrative creation "sandpit" environment.

We can consider both *analysis*, i.e. breaking up of text units, and *synthesis*, i.e. assembling texts into a larger text unit.

In Arisotle (350 BC) the component parts of a play, for example, are considered. Under the title of "Outlines and episodization", it holds, according to Aristotle: "Stories … should first be set out in universal terms … on that basis, one should then turn the story into episodes and elaborate it." He continues: "… reasoning is the speech which the agents use to argue a case or put forward an opinion". It is interesting how the decomposition into episodes is related to agents who provide sense and meaning to the component parts. This is a theme which we will return to below. It may be noted, cf. White (2001), that Aristotle's perspectives on story and narrative retain importance to this day.

In narrativization we seek to build a story-line from an arbitrarily large number of texts. We can achieve a fixed target story-line size or length. There is traceability of all component parts of the story-line (which, we may note, is potentially important for propagation and preservation of digital rights). By integrating story transitions with heuristics for structuring story we allow additional story sub-plot embedding, and the placement of other relevant information. Furthermore, by appropriate mapping of story transitions, we can allow interactive reading and other adaptive usage frameworks.

## 3. Role of Machine Learning and Data Mining in Filmscript Analysis

We have initially chosen to look at filmscript as a particular kind of text, both as a text in itself which is broken up into discrete units, with considerable amounts of easily recognizable metadata attached to each scene, but also as a text which is intended to achieve itself in a different visual form which will be experienced chronologically and in real time.

A filmscript, expressing a story, is the starting point for any possible production for cinema or TV. See the short extract in Figure 1. TV episodes in the same series may each be developed by different scriptwriters, and later by different producers and directors. The aim of any TV screenplay is to provide a unique but repeatable experience in which each episode shares certain structural and narrative traits with other episodes from the same series despite the fact they may have been originated or realized by different people or teams. There is a productive tension between the separate needs for uniqueness, such that each episode seems fresh and surprising, but also belongs to its genre. For example an episode of Friends needs to feel Friends-like, to offer the specific kinds of pleasure the audience associates with the series. We propose that these distinctive qualities of any individual script and the distinctive qualities of any genre are open to analysis through a pattern and trend analysis toolset that finds distinctive ways of representing the essential structural qualities of any script, and the series to which it belongs and thus enables the writer, the script developer or producer to have a deeper understanding of the script and have objective criteria for the creative decisions they take. Moreover as the scripts are migrated to digital formats the tools offer many possibilities for prototyping from the information gathered.

By analysis of multiple screenplays, TV episodes and genres the technology will allow the possibility of creating distinctive analytical patterns for the structure of genres, series, or episodes in the same way that comparative authorship can be assessed for individual writers. Large numbers of filmscripts, for all genres, are available and openly accessible online (e.g. IMSDb, The Internet Movie Script Database, www.imsdb.com; or www.kilohoku.com).

A filmscript is a semi-structured textual document in that it is subdivided into scenes and sometimes other structural units. Both character dialog and also descriptive and background metadata is provided in a filmscript. The metadata is partially formalized since there are some invariants in any filmscript including set of characters; and essential metadata related to location (internal, external; particular or general location name); characters; day, night. Accompanying the dialog there is often description of action in free text. While offering just one data modality, viz. text, there is close linkage to other modalities, – visual, speech and sometimes music.

```
[INT. CSI - EVIDENCE ROOM -- NIGHT]

(WARRICK opens the evidence package and takes out the shoe.)

(He sits down and examines the shoe.  After several dissolves,
WARRICK opens the lip of the shoe and looks inside.  He finds something.)

WARRICK BROWN:  Well, I'll be damned.

(He tips the shoe over and a piece of toe nail falls out onto
the table.  He picks it up.)

WARRICK BROWN:  Tripped over a rattle, my ass.
```

Figure 1: Example of a scene from a script.  This is a short scene, scene 25, from the CSI (Crime Series Investigation, Las Vegas) TV series.  This is the very first, 1X01, Pilot, orginal air date on CBS Oct. 6, 2000.  Written by A.E. Zuiker, directed by D. Cannon.  Script available in full from TWIZ TV (Free TV Scripts & Movie Screenplays Archives), twiztv.com .

A substantial part of our work is focused on data mining, understood as unsupervised learning or knowledge discovery.

It is interesting to note that another important line of enquiry has been pursued in the supervised or machine learning direction.  The feasibility of using statistical learning methods in order to map a feature space characterization of film scripts (using two dozen characteristics) onto box office profitability was pursued by Eliashberg et al. (2007).  The importance of such machine learning of what constitutes a good quality and/or potentially profitable film script has been the basis for (successful) commercial initiatives, as described by Gladwell (2006).  The business side of the movie business is elaborated on in some depth in Eliashberg et al. (2006).

Our aim in our work is to complement the machine learning work of Eliashberg and his colleagues at the Wharton School, University of Pennsylvania, with our data mining approaches.  The machine learning approach attempts to map filmscript inputs to target outputs (in practice, there is considerable manual intervention involved in data preprocessing).  An important part of our data mining approach attempts to find faint patterns and structure in the filmscript data that can be exploited for a range of decision making, including such learning of mapping onto box-office success.

In Murtagh et al. (2009a), we carried out a detailed study of features proposed and discussed by McKee (1999).  Analysis of style and structure was carried out, using 999 randomizations of scenes thereby furnishing what is needed for testing the statistical hypothesis of significance at the 0.001 level, of the sequence of 77 scenes in the movie Casablanca. In all, 9 attributes were used. Through the randomizations of scene sequence we have a Monte Carlo approach to test statistical significance of the given script's patterns and structures as opposed to randomized alternatives.

We found the following, in summary, in this Monte Carlo "baselining" analysis. The entire Casablanca plot is well-characterized by the variability of movement from one scene to the next (attribute 2). Variability of movement from one beat to the next is smaller than randomized alternatives in 82% of cases. Similarity of orientation from one scene to the next (attribute 3) is very tight, i.e. smaller than randomized alternatives. We found this to hold in 95% of cases. The variability of orientations (attribute 4) was also tighter, in 82% of cases. Attribute 6, the mean of ups and downs of tempos is also revealing. In 96% of cases, it was smaller in the real Casablanca, as opposed to the randomized alternatives. This points to the "balance" of up and down movement in pace.

### 4. An Overview of Computational Technology Applied

A primary tool that we use is Correspondence Analysis (a mathematical data analysis method), allied with automatic classification, to discover deep structure in narrative and also to explore the relationship between textual prototypes of visual narratives and the narrative itself. We base such analysis on the raw text – the set of all words used – possibly preserving aspects of its original sequence.

For McKee (1999) text is the "sensory surface of a work of art" and reflects the underlying emotion or perception. Based on this, Murtagh et al. (2009a) consider the following approach to textual data mining, with a range of novel elements. Firstly, the orientation of narrative through Correspondence Analysis (Benzécri, 1979; Murtagh, 2005; Bourdieu, 2010) maps scenes (and subscenes), and words used, in a near fully automated way, into a Euclidean space representing all pairwise interrelationships. Such a space is ideal for visualization. Interrelationships between scenes are captured and displayed, as well as interrelationships between words, and mutually between scenes and words. So in Murtagh et al. (2009a) we discerned story semantics arising out of orientation of narrative. See illustration in Figure 2.

Secondly, we examined caesuras and breakpoints, by taking the Euclidean embedding further and inducing an ultrametric on the sequence of scenes. In Murtagh et al. (2009) we related these results to visualization. These proof of concept studies used the filmscript for Casablanca (Burnett and Allison, 1940) and a range of CSI episodes.

A succinct view of our data analysis methodology is as follows.

Initial data are in the form of frequencies of occurrence, which can be in the simple form of presence/absence. We start therefore with, e.g., scenes crossed by words. Typically we use all words, with the only limitations being that isolated single character words are ignored, punctuation is replaced by a blank character, and upper case is converted to lower case. While there is some loss of information in this selection of words, nonetheless our approach uses all remaining words, and is fully automated. This differs from the usual current approach which avails of lemmatization and perhaps even grammatical preprocessing of text. We prefer to use all words, including "tool words" ("to", "the", "and", and so on; these are words that are usually put on stop lists in order to be removed from consideration) precisely because these tool words not only express style but often semantics as well. This is discussed, with published examples, in Murtagh (2005: 166-172).

On the frequency of occurrence data, we define a chi squared metric (see Murtagh, 2005). This metric entails a weighting of rows (e.g. scenes) and columns (e.g. words) to account for relative proportions. Mapping is then carried out to furnish a factor space that is based on the Euclidean metric: see Figure 2 for a best planar projection of this Euclidean space. Finally from the latter Euclidean space a hierarchical clustering is induced, which defines an ultrametric topology. We remark therefore that our data analysis pipeline is "a tale of three metrics" – chi squared, Euclidean and ultrametric.

In Murtagh et al. (2009a) we explored the semantics of narrative. We tracked emotional change over time; we developed and statistically appraised various semantic feature sets; and we looked at change and novelty. The filmscript of Casablanca was used for this. In Murtagh et al. (2009) television film was studied, using the CSI (Crime Scene Investigation) series. Characterization of thematic change and novelty were at issue.

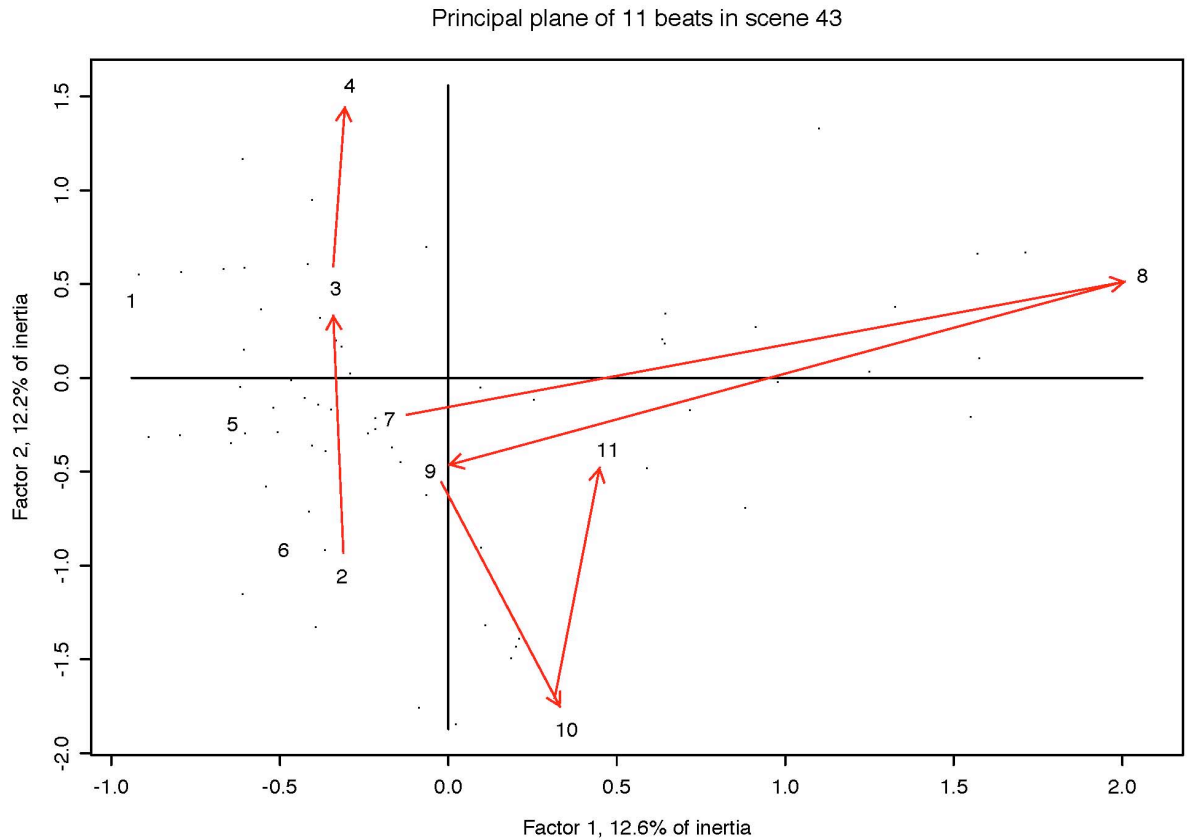

Figure 2: Analysis of 11 subscenes, called "beats", in scene 43 of the film Casablanca. This scene relates to Rick and Ilsa in the marketplace seeking black market visas. Rapprochement of Rick and Ilsa is expressed by downward movement from beat to beat on the ordinate. In beat 8, Ilsa indicates surprisingly that she will leave Casablanca. See Murtagh et al., 2009a, for full discussion.

Semantic analysis for us is the study of all possible pairwise relationships, provided by the Correspondence Analysis (cf. the planar projection in Figure 2); and the study of change on all possible scales provided by the hierarchical clustering (see Figure 3).

Figure 3 is from the analysis of the popular television series, CSI Las Vegas, and relates to the pilot episode, CSI 101, consisting of 50 scenes, and 1679 unique two or more character words. It is interesting to note that the pre-planned commercial breaks were between scenes 4 and 5; between 14 and 15; then 25 and 26; and finally between 32 and 33. This hierarchical clustering method takes account of the given sequence of scenes. Further discussion can be found in Murtagh et al. (2009b).

Recent work on segmentation, or episodization, based on sequence-constrained hierarchical clustering, has included a comparative study of multiple versions of a traditional ballad. See Murtagh and Ganz (2010).

The aims of the work described in this section have included analysis of deep structure in filmscript: identity, drivers of attention and story evolution. At an overall level, and by extension, our work can be considered too in terms of scientific application to creativity.

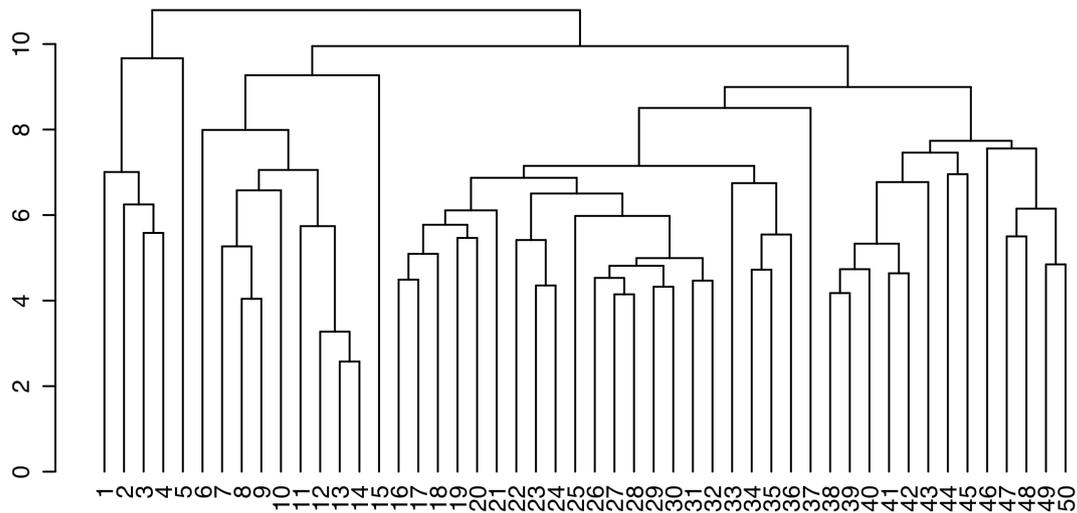

Figure 3: Sequence-constrained complete link hierarchical clustering of the CSI 101 television program. Scene 15 is seen to be quite different from the immediate preceding, and immediate following scenes. So too, scene 37 is quite anomalous.

5. Integration and Assembling of Narrative

Narrativization is how we term the finding and tracking of the narrative. The tracking of narrative has been at issue in the Casablanca work described above, where we tracked emotional content. Tracking can be pattern finding at a particular time (or scene or chapter) or it can be the labeling of change over time. Analysis of change over time has been a central concern in the CSI television script analysis. We now look at the synthesis task, moving on from the analysis task.

An approach was developed for a support environment for distributed literary creation. Author Reddington used this support environment in the TooManyCooks Project in which a set of student volunteers from RHUL's Arts Faculty (see Figure 4) collectively wrote a novel in a week. This was printed and self-published as a book (Cooks, 2009a), where the author name is a collective (virtual) one. A second and separate project of the same sort resulted in another collectively created work, a book composed in one week (Cooks, 2009b).

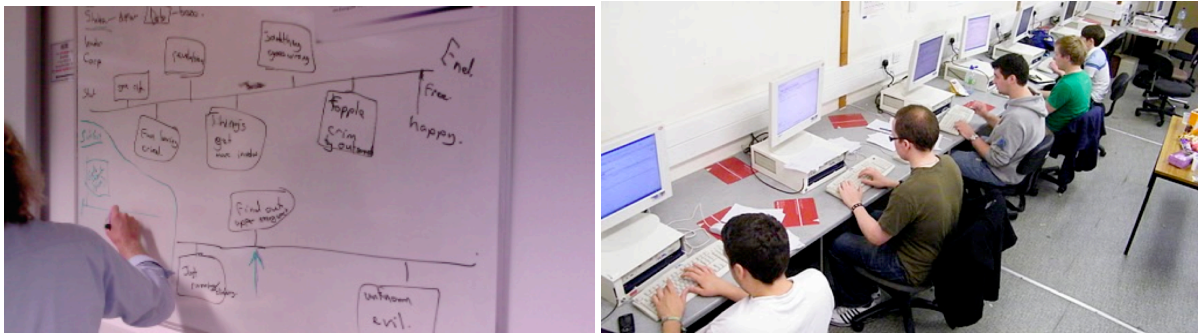

Figure 4: Work underway that resulted in Cooks (2009a, 2009b).

The support environment here was for collaborative, distributed creating of narrative. This includes the following: pinpointing anomalous sections; assessing homogeneity of style over successive iterations of the work; scenario experimentation and planning; condensing parts, or elaborating; and evaluating similarity of structure relative to best practice in the chosen genre.

Our work on narrativization is relevant for general information spaces, including all forms of human-machine interaction.

A narrative functions effectively because in simple terms it suggests a causal or emotional relationship between events. A story is an expression of causality or connection. Its aim is the transformation of a number of discrete elements (facts or views or other units of information) into part of a composite – the narrative. This process also involves rejecting those elements that are not germane to that narrative. Thus a narrative simultaneously binds together those elements that are germane to it and, through a process of editing and retelling, discloses those that are not part of that particular narrative. This process is discussed in Bruner (1991).

We have been exploring how our algorithms can be used to structure user generated content "narratively". This includes, firstly, the ordering or reordering of contributions. We want these contributions to best represent a narrative that can be read from the totality of contributions to any particular thread of activity (including e.g. comments on news stories or posts in online forums). Secondly, we seek to discard or render invisible contributions that do not take the narrative forward. Any set of contributions to a thread are to be arranged beyond chronological and thematic structuring to create narrative, using salience and interestingness detection. Finally, we look at those contributions that change the future contributions to a thread (just as our algorithms have been exploited to detect the turning points and act breaks in a screenplay).

In Theune et al. (2002), an algorithmic, distributed narrative story engine is described. Modules are responsible for plot, narrative and presentation. However plot and narrative are interlinked in an intricate way. So in our work we want this inherent relationship to be maintained. Implementation-wise, we avail of an interactive support environment. We can essentially say that in our approach the semantics can be traced out, or demarcated, through the syntax elements, and reciprocally (not least through our use of "tool words") the semantics can be manifested above and beyond the syntax. We can note therefore how, for us, plot, narrative and presentation are all interwoven, and how we address this.

## 6. Exploiting Typicality and Atypicality for Narrative Summarization and Transcoding

Taking Marx's account of commodity fetishism (Marx, 1867), we subdivided it into 21 consecutive paragraphs. Word counts varied from 512 words in paragraph 6 to just 25 words in the one-sentence paragraph 5.

Figure 5 using all the data gives rise to the following. Most paragraphs are bunched near the origin, i.e. the average paragraph profile. (One speaks of "profiles" in Correspondence Analysis because paragraphs – rows – are normalized in the analysis to even out the different numbers of terms per paragraph. Similarly term – column – profiles arise through the same evening out of the different numbers of paragraphs characterized by each term.)

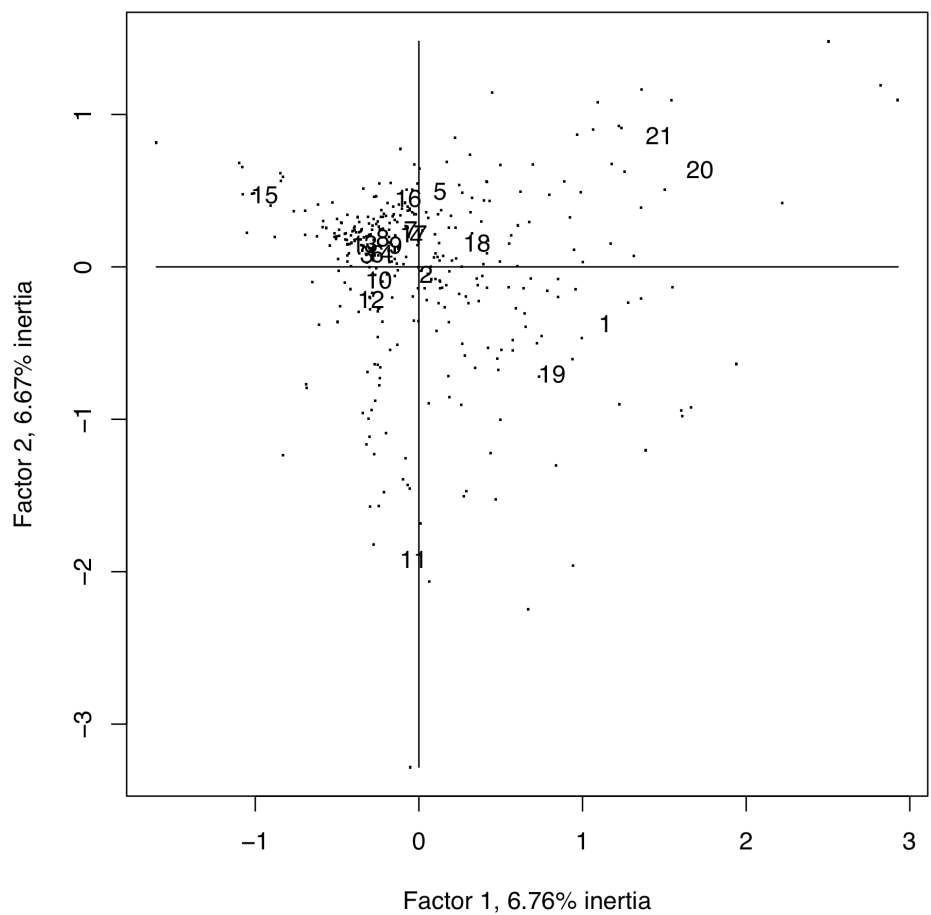

Figure 5: Projections on this planar visualization of paragraphs 1 to 21 are noted. The 974 terms that characterize these paragraphs are displayed as dots. Percentage inertia quantifies the relative information of the factor or axis.

So the paragraphs that seem to really matter are: 15, 21, 20, 1, 19, and 11. By "really matter" we can quantify this through the mathematically defined correlations and contributions to the factors.

As a consequence of these findings in Figure 5, we will proceed with just the restricted set of 6 paragraphs as constituting the "backbone" of Marx's narrative.

We find that Figure 6, using 6 paragraphs instead of the given 21, and hence with 482 associated terms rather than 974 associated terms, is not unlike Figure 5. It is important to note that the factors of Correspondence Analysis are not fixed in their orientations. Hence, it is equally acceptable to have reflections in the axes, just so long as there is consistency. So in both figures, paragraphs 20 and 21 are closely located. These two paragraphs are counterposed to paragraph 11. In both figures, 20 and 21, and then 11 and finally 15 constitute approximately three vertices of a triangle. What does change in location in proceeding to Figure 6 from Figure 5 is paragraph 19; and paragraph 1 has come closer to the origin (and hence the average paragraph).

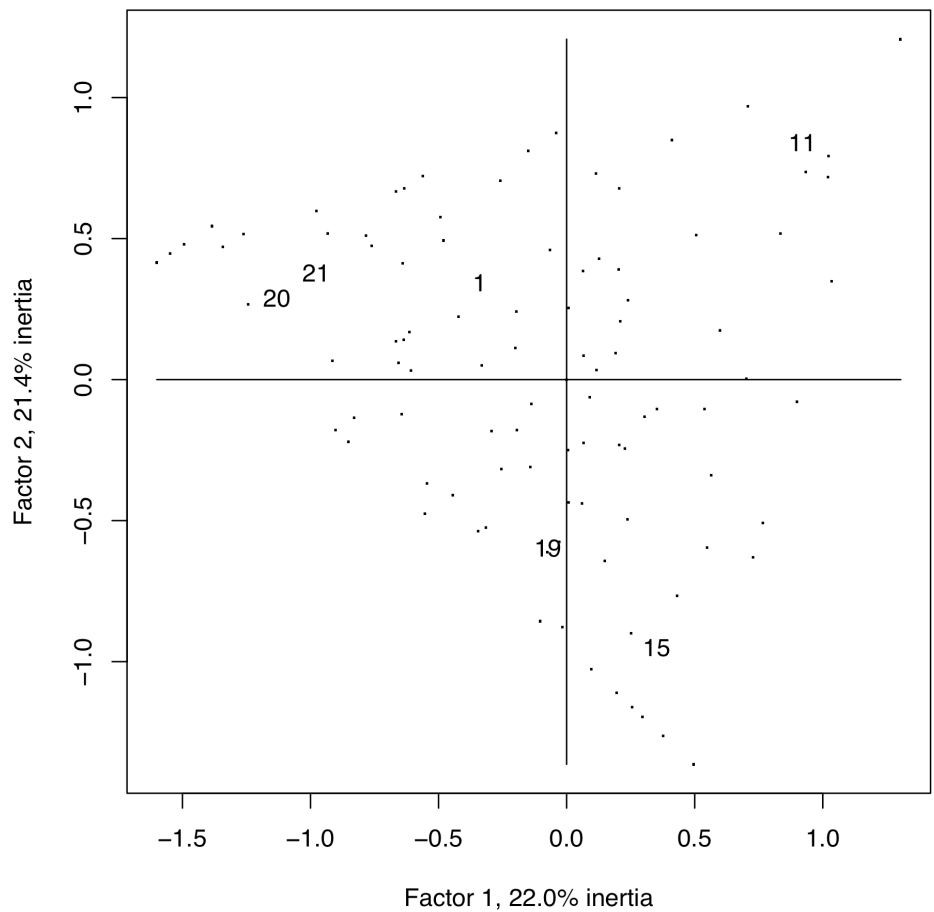

Figure 6: Compared to Figure 5, a smaller set of paragraphs is used here: paragraphs 1, 11, 15, 19, 20, 21. There are 482 terms in these paragraphs, displayed as dots.

If one wished to have the most salient paragraphs to summarize the overall text we therefore find this job to be done quite well by using paragraphs 1, 11, 15, 19, 20 and 21. The reader of the text on commodity fetishism who is in a hurry should concentrate on these paragraphs!

In brief, the contents of the paragraphs are as follows.

Paragraph 1: commodity's attributes
Paragraph 11: Robinson Crusoe story
Paragraph 15: religion and worship
Paragraph 19: on provenance of monetary system
Paragraph 20, 21: use value and/or exchange value

Marx's writing can be very visual, with plenty of illustrations. For instance, in paragraph 1, there is reference to tables made from wood.

What we have exemplified in this case study is semantically-based narrative summarization. Apart from summarization, this analysis approach can be used also for selecting core visual metaphors or other features for multiplatform transcoding.

As regards visual metaphors needed in taking the script into the film-set, or performance environment, text in time is designed to create visual metaphors in the teller and the audience as Ganz (2010) has explored in writing about the links between oral narrative and the cinema. Ganz (2011), in a chapter in a forthcoming book on *Storytelling in World Cinema*, considered how it is the dynamic invocation of images that is an essential element of oral narrative. These ideas are also discussed in the work of Elizabeth Minchin: "The storyteller's language serves as a prompt and a guide to stimulate us to perform the exercise of visualization and to ensure that the picture which we build up is appropriate."

**7. Conclusion: Between Film and Game**

Filmscript, expressing a story, is hugely important in its own right. Elaborations and extensions of this context include automating the film production process, or studying visual semantics. We will extend our work in such directions in the future. It is clear though that we will be informed by indirect linkage with film, on many levels, since this constitutes the targeted end product of any filmscript. McKee (1999) bears out the great importance of the script: "50% of what we understand comes from watching it being said." And: "A screenplay waits for the camera. … Ninety percent of all verbal expression has no filmic equivalent."

Our work is relevant for areas above and beyond filmscript. Filmscripts are a semi-formalized way of representing a story. Filmscripts are a way to capture in a partially structured way the narrative or storyline. So filmscripts make it easier to explore and understand a story, indeed to unravel and analyze the semantics of the narrative (Murtagh et al., 2009a). The convergence of story-telling and narrative, on the one hand, and games, on the other, is explored in Glassner (2004). There and in Riedl and Young (2006), story graphs and narrative trees are used as a way to open up to the user the range of branching possibilities available in the story. Among other important new research directions that we have opened up is the art, and the engineering, involved in turning a TV series into material for games.

(30) E. White, "Aristotle, drama and the craft of reality TV", Television Review, online Media Life Magazine, 22 March 2001. http://www.medialifemagazine.com/news2001/index.html .